\documentclass{article}

\usepackage{arxiv}

\usepackage[utf8]{inputenc} % allow utf-8 input
\usepackage[T1]{fontenc}    % use 8-bit T1 fonts

\usepackage{url}            % simple URL typesetting
\usepackage{booktabs}       % professional-quality tables
\usepackage{amsfonts}       % blackboard math symbols
\usepackage{nicefrac}       % compact symbols for 1/2, etc.
\usepackage{microtype}      % microtypography
\usepackage{lipsum}		% Can be removed after putting your text content
\usepackage{graphicx}
\usepackage[numbers,square]{natbib}
\usepackage{hyperref}       % hyperlinks
\usepackage{doi}
\usepackage{amsmath} 
\usepackage{float}

\title{Emotion Recognition and Generation: A Comprehensive Review of Face, Speech, and Text Modalities}

\author{ \href{https://orcid.org/0009-0005-9159-6528}{\includegraphics[scale=0.06]{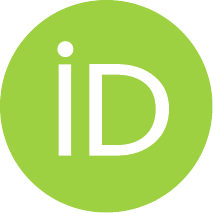}\hspace{1mm}Rebecca Mobbs} \\
	School of Computer Science and Mathematics\\
	Kingston University\\
	London \\
	\texttt{k2369889@kingston.ac.uk} \\
	\And
	\href{https://orcid.org/0000-0001-6170-0236}{\includegraphics[scale=0.06]{orcid.pdf}\hspace{1mm}Dimitrios Makris} \\
	School of Computer Science and Mathematics\\
	Kingston University\\
	London \\
	\texttt{d.makris@kingston.ac.uk} \\
    	\And
	\href{https://orcid.org/0000-0003-4679-8049}{\includegraphics[scale=0.06]{orcid.pdf}\hspace{1mm}Vasileios Argyriou} \\
	School of Computer Science and Mathematics\\
	Kingston University\\
	London \\
	\texttt{vasileios.argyriou@kingston.ac.uk} \\
}

\hypersetup{
pdftitle={A template for the arxiv style},
pdfsubject={q-bio.NC, q-bio.QM},
pdfauthor={David S.~Hippocampus, Elias D.~Striatum},
pdfkeywords={First keyword, Second keyword, More},
}

\begin{document}
\maketitle

\begin{abstract}
Emotion recognition and generation have emerged as crucial topics in Artificial Intelligence research, playing a significant role in enhancing human-computer interaction within healthcare, customer service, and other fields. Although several reviews have been conducted on emotion recognition and generation as separate entities, many of these works are either fragmented or limited to specific methodologies, lacking a comprehensive overview of recent developments and trends across different modalities. In this survey, we provide a holistic review aimed at researchers beginning their exploration in emotion recognition and generation. We introduce the fundamental principles underlying emotion recognition and generation across facial, vocal, and textual modalities. This work categorises recent state-of-the-art research into distinct technical approaches and explains the theoretical foundations and motivations behind these methodologies, offering a clearer understanding of their application. Moreover, we discuss evaluation metrics, comparative analyses, and current limitations, shedding light on the challenges faced by researchers in the field. Finally, we propose future research directions to address these challenges and encourage further exploration into developing robust, effective, and ethically responsible emotion recognition and generation systems. 
\end{abstract}

% keywords can be removed

\section{Introduction}

Emotions are central to human communication, shaping interactions through body language, facial expressions, vocal intonations, and textual cues \cite{tian2001recognizing}. Psychological research suggests recognition of emotions is innate in humans, with newborns able to replicate facial expressions and vocal tones as early as two days old \cite{johnson2021face}. Understanding emotions aids in teamwork and cooperation, a concept recognised by Darwin's theories on survival mechanisms \cite{darwin1998expression}. This significance has led to the development of emotion models like Ekman and Friesen’s Facial Action Coding System (FACS), which categorises emotions such as anger, disgust, fear, happiness, sadness, surprise, and contempt \cite{ekman1971constants,ekman1994strong}, forming the basis for many contemporary emotion recognition systems.

As interest in artificial intelligence (AI) grows, emotion recognition and generation technologies have gained traction in fields such as healthcare, customer service, education, and entertainment \cite{einfochips2024website,elevateai2024website,brandt2024facial,huang2023mental,ramis2020using,tang2020classroom}. AI systems can now analyse and simulate emotional responses, allowing machines to engage in more meaningful human-computer interactions. Emotion recognition is used in applications such as driver fatigue detection \cite{liu2020driver} and lie detection \cite{rizwan2020accurate}, while generative models create realistic emotional content in apps like FaceApp \cite{sansar2023societies}, HeadSpace \cite{wasil2022there}, and Wysa \cite{beatty2022evaluating}.

This survey provides a comprehensive review of State-of-the-Art (SOTA) methodologies in AI for emotion recognition and emotion generation, addressing the gap in the literature regarding the integration of these two domains and their applications across multiple modalities. The generation of emotions on faces, Facial Expression Generation (FEG) systems, are termed in the literature as Talking Face or Speech/Text-to-Animation models, while Speech Emotion Generation (SEG) involves Speech-to-Speech or Speech Reenactment methods, and Text Sentiment Generation (TSG) relies on Large Language Models (LLMs). Existing reviews have typically focused on either emotion recognition \cite{Adyapady2023,Deng2023,al2023speech} or emotion generation \cite{cao2023comprehensive,harshvardhan2020comprehensive}, without addressing their intersection. Additionally, Facial Expression Recognition (FER) and FEG have not yet been discussed alongside Speech Emotion Recognition (SER), SEG, or Text Sentiment Recognition (TSR). Research tends to prioritise facial systems due to heightened public interest and the relative ease with which facial expressions are interpreted by both humans and machines \cite{li_2020}. These systems also benefit from extensive pretrained models and datasets derived from computer vision research \cite{toisoul2021estimation}. By exploring both emotion recognition and generation across modalities, this survey aims to offer insights into current techniques, highlight areas for improvement, and guide future research directions.

This survey is structured to provide a holistic examination of the field. Section 1.1 explores various applications of emotion recognition and generation models. Section 2 discusses preprocessing techniques to improve model accuracy and efficiency. Section 3 reviews the datasets commonly used, detailing their characteristics. Sections 4 and 5 present state-of-the-art methods for emotion recognition and generation, respectively, across faces, speech, and text. Section 5.4 discusses emotion control methods accross modalities. Section 6 provides a comparative analysis of evaluation metrics to assess SOTA performance. Section 7 outlines current challenges and future research directions. Finally, Section 9 concludes with a synthesis of key findings and contributions to the development of emotion recognition and generation technologies.

\subsection{Applications} %mention these systems would be good for use with people with hearing or speech impairments
Emotion recognition systems are used across various fields. In customer service, they are utilised to discern customers' emotions and evaluate the effectiveness of sales assistants' communication strategies through assessment of transcripts \cite{elevateai2024website}. Similarly, at self-service checkouts, FER is used to gauge customer satisfaction based on their facial cues \cite{einfochips2024website}. In healthcare, these systems assist in tracking the progression of Alzheimer’s disease \cite{brandt2024facial}, facilitating therapy sessions \cite{huang2023mental}, and supporting individuals with Asperger’s Syndrome in recognising emotions \cite{banerjee2023training}. They are also used in robotics to interpret human emotions during interactions with machines \cite{ramis2020using}, and in educational settings to evaluate students' engagement and learning \cite{tang2020classroom}. Other applications include lie detection \cite{rizwan2020accurate} and monitoring driver fatigue levels \cite{liu2020driver}.

Emotion recognition systems can also serve as foundational tools for training models capable of generating realistic emotional content \cite{li_2020}. These models can be used to create visual virtual assistants and avatars for virtual calls \cite{yasuoka2022effects}. As reliance on chatbots for social interactions and advice increases \cite{van2021m}, there is a growing opportunity for the development of talking head chatbots. Such chatbots would use speech or text input—whether from a customer service representative, therapist, or a text generation model—to produce animated faces with lifelike emotions in real-time. These animated avatars could integrate with AI models such as Character.AI \cite{characterAI}, ChatGPT \cite{achiam2023gpt}, Llama \cite{llama}, or Gemini \cite{team2023gemini} to function as therapeutic or customer service bots. This technology has the potential to provide users with a highly immersive and personalised experience, enhancing or even replacing current customer service chatbots. 

\section{Preprocessing for ER and EG Systems}
Preprocessing is an important stage in deep learning pipelines, particularly when handling data obtained from uncontrolled or 'in-the-wild' environments, such as facial and speech data extracted from movies or textual data from social media. Such data often exhibit significant variability compared to controlled laboratory settings, with variations in background, lighting, noise, and other artefacts. To address these challenges, preprocessing typically involves standard steps like data normalisation, noise reduction, and feature extraction to ensure data consistency and optimise model performance. Below, we explore the specific preprocessing techniques used for processing face, speech, and textual data.

\subsection{Preprocessing for Face Systems}
Preprocessing for facial emotion recognition systems  aims to enhance image quality, standardise data, and extract critical features for accurate model predictions. The initial step involves resizing and cropping facial images to create uniform input dimensions, ensuring consistency across the dataset. By eliminating background elements and focusing on the region of interest, these techniques enable models to concentrate on key facial features. Normalisation, through scaling pixel values to a common range (e.g., 0 to 1 or -1 to 1), ensures uniform pixel intensity across different samples, thereby enhancing the model’s capacity to learn relevant patterns. Common methods such as mean subtraction \cite{Krizhevsky2012} and standard deviation normalisation \cite{LeCun1998} are frequently used. Noise reduction techniques, like Gaussian blurring \cite{Gonzalez2002} and median filtering \cite{Tukey1977}, are used to minimise the impact of noise introduced during image acquisition or transmission.

Techniques such as histogram equalisation \cite{Pizer1987} improve contrast by redistributing pixel intensities, enhancing visibility in images captured under challenging conditions. Data augmentation, involving transformations like rotation, scaling, and flipping, increases training data diversity and mitigates overfitting \cite{Shorten2019}. Furthermore, advanced algorithms such as Haar cascades \cite{Viola2001} and deep learning-based facial landmark detection methods \cite{Zhang2014} are applied to extract and align facial regions, standardising poses and reducing variability. Feature extraction models, such as VGG \cite{Simonyan2014}, ResNet \cite{He2016}, and MobileNet \cite{Howard2017}, are widely used for extracting high-level features. Colour space transformations and quality control measures help streamline data preparation, ensuring only high-quality data is fed into the models \cite{Gonzalez2002, Wang2004}.

\subsection{Preprocessing for Speech Systems}
The primary goals of preprocessing in speech systems are noise reduction, normalisation, segmentation, and feature extraction from raw audio signals. Noise reduction methods like spectral subtraction \cite{Boll1979}, Wiener filtering \cite{Lim1979}, and adaptive filtering \cite{Widrow1985} are used to eliminate background noise which can degrade speech signal quality. Normalisation adjusts amplitude and dynamic range to maintain consistency across recordings \cite{Rabiner1993}. Speech segmentation techniques, such as endpoint detection \cite{Rabiner1975} and silence removal \cite{Sadjadi2013}, isolate speech segments within continuous audio streams, enabling more targeted analysis.

Feature extraction captures the salient characteristics of speech, using Mel-Frequency Cepstral Coefficients (MFCCs) \cite{Davis1980}, which represent spectral properties in a compact form, and Linear Predictive Coding (LPC) \cite{Makhoul1975}, which models the spectral envelope. Other methods like pitch estimation \cite{Boersma1993} and anti-aliasing filtering \cite{Crochiere1983} help preserve signal integrity. Techniques such as de-reverberation \cite{naylor2010speech} and pre-emphasis \cite{o1988speech} further refine the signal quality. For segmentation, windowing techniques like frame blocking divide speech signals into shorter frames, facilitating computational efficiency \cite{hamid2018frame}. Mean and variance normalisation standardises feature scales, improving model robustness to variability in input data \cite{viikki1998recursive}.

\subsection{Preprocessing for Text Systems}
Text preprocessing begins with tokenisation, which breaks down text into smaller units, such as words or characters. This is followed by lowercasing, which standardises the text by treating uppercase and lowercase versions of words identically, thereby reducing vocabulary size and simplifying the learning process \cite{Mikolov2013}. Punctuation and special character removal further eliminate noise which could interfere with learning. Stopwords—such as “and” or “the”—are often removed, as they carry little semantic value \cite{Manning2008}. Stemming and lemmatisation techniques group words with similar meanings, helping models understand linguistic variations \cite{Porter1980, Bird2009}.

% \clearpage % Ensures everything before this point is completed on previous pages
% \input{datasets_table}
% \clearpage % Ensures that the content after this point starts on a new page

Numerical values are encoded or replaced with placeholders to maintain the semantic integrity of the text \cite{Ghosh1994}. Out-of-vocabulary words are managed through tokenisation or character-level representations \cite{Pennington2014}, while padding and truncation ensure uniform sequence lengths, which is crucial for text classification \cite{Goldberg2017}. Pretrained word embeddings, such as Word2Vec \cite{Mikolov2013_2}, can be used to initialise the embedding layers of deep learning models or be fine-tuned during training. Encoding methods like one-hot or integer encoding convert textual data into numerical representations, while pretrained tokenisers accelerate this conversion \cite{Johnson2017}. Text augmentation techniques, such as synonym replacement and paraphrasing, diversify training data and reduce overfitting, improving generalisation \cite{Wei2019}.

\section{Datasets for Face, Text, and Speech ER and EG Systems}
High-quality, diverse datasets are essential for training emotion recognition and generation models. These datasets provide labelled examples from facial expressions, speech, and text, enabling models to learn emotional cues in varied contexts. Some datasets are captured in controlled environments, while others are collected in the wild, offering more complex real-world variations. This section highlights the most widely used datasets across facial, speech, and text systems, focusing on those with comprehensive emotional labelling and diversity (see \ref{tab:datasets}). 

\begin{table}[h]
\small
\centering
\begin{tabular}{|p{3cm}|p{5cm}|p{2cm}|p{2cm}|p{3cm}|}
\hline
\textbf{Name} & \textbf{Description} & \textbf{Type} & \textbf{Size} & \textbf{Emotions} \\ \hline

AffectNet & Extensive facial imagery dataset annotated with discrete and continuous emotion labels. & Image & 450,000 images & Surprise, fear, disgust, happiness, sadness, anger, neutral, contempt \\ \hline

RAF-DB & Diverse facial expression dataset featuring multiple genders, ages, and ethnicities. & Image & 29,672 images & Surprise, fear, disgust, happiness, sadness, anger, neutral \\ \hline

FERPlus & Derived from the FER2013 dataset, enhancing expression annotations through crowdsourcing. & Image & Unlimited & Surprise, fear, disgust, happiness, sadness, anger, neutral, contempt \\ \hline

AFEW & High-resolution videos from YouTube with over 300 subjects and 10,000 sentences. & Video & 16 hours & Surprise, fear, disgust, happiness, sadness, anger, neutral \\ \hline

HDTF & Video clips gathered from TV shows and movies, including various head poses and occlusions. & Video & 1,809 clips & Surprise, fear, disgust, happiness, sadness, anger, neutral \\ \hline

AFEW-VA & Video clips annotated for valence and arousal levels, with 68 facial landmarks per frame. & Video & 600 clips & Surprise, fear, disgust, happiness, sadness, anger, neutral \\ \hline

DFEW & Facial expression dataset created from more than 1,500 movies. & Video & 12,059 clips & Happiness, anger, sadness, fear, disgust, surprise, neutral \\ \hline

CK+ & Laboratory-controlled video data capturing transitions from neutral to peak expression. & Video & 593 sequences & Surprise, fear, disgust, happiness, sadness, anger, contempt \\ \hline

MEAD & High-resolution emotional audiovisual dataset with 60 actors. & Video \& audio & 16,800 hours & Surprise, fear, disgust, happiness, sadness, anger, contempt \\ \hline

LRW & Video sequences of people speaking words in uncontrolled conditions. & Video \& audio & 1,000 utterances & Unlabeled \\ \hline

LibriTTS & Multi-speaker English corpus of read speech at 24kHz for TTS research. & Audio & 585 hours & Unlabeled \\ \hline

VCC2018 & Dataset for speech-to-speech systems, consisting of male and female speakers. & Audio & 464 sentences & Unlabeled \\ \hline

ESD & Collection of audio recordings for studying emotions expressed through speech. & Audio & 7,000 utterances & Neutral, happy, angry, sad, surprise \\ \hline

Empathetic Dialogues & Open-domain conversations between speakers and listeners for empathic responses. & Audio & 24,850 conversations & 32 emotion labels \\ \hline

EMO-DB & German emotional speech recorded by ten professional speakers. & Audio & 535 utterances & 7 emotions \\ \hline

CASIA & Mandarin emotional speech dataset. & Audio & 1,200 snippets & 6 emotions \\ \hline

Amazon Reviews & Large dataset of product reviews provided by Amazon. & Text & Unlimited & - \\ \hline

Twitter & Collection of tweets for social media text analysis. & Text & Unlimited & - \\ \hline

Reddit & Comments and posts from Reddit for understanding informal language. & Text & Unlimited & - \\ \hline

\end{tabular}
\caption{Datasets for ER and EG Systems}
\label{tab:datasets}
\end{table}

\section{Emotion Recognition for Faces, Speech, and Text}
\begin{figure}[hbt!] 

     \centering 

     \includegraphics[width=\textwidth]{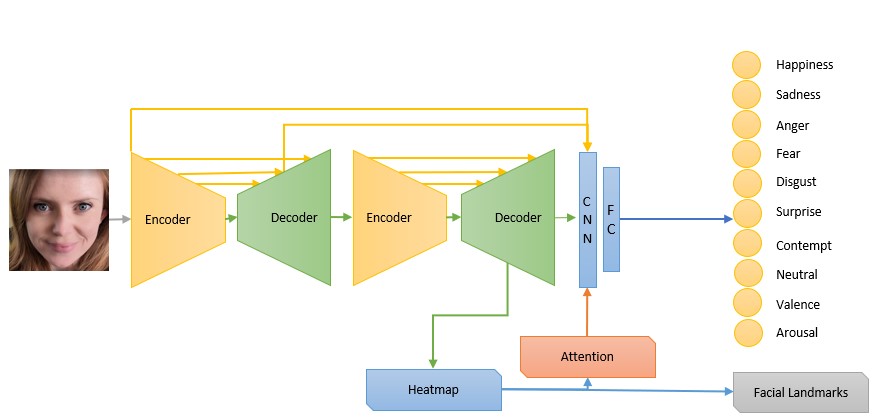} 

     \caption{The EmoFAN pipeline integrates facial landmark detection, discrete emotion classification, and continuous valence-arousal estimation in a single neural network. This unified model performs all tasks in one pass, using a face-alignment network and an attention mechanism to focus on key facial regions, enhancing accuracy. Joint prediction of both emotion types, combined with knowledge distillation, improves robustness.\cite{toisoul2021estimation}} 

     \label{emofan} 

 \end{figure}

This section will discuss deep learning methodologies for emotion recognition for faces, speech, and text. We will discuss the strengths and limitations of current literature. Most emotion recognition systems use the 8 primary emotions anger, disgust, fear, happiness, sadness, surprise, contempt, and neutral \cite{ekman1994strong}. Unlike traditional methodologies where feature extraction and classification are treated as distinct stages \cite{Schuller2010}, deep learning frameworks for emotion detection enable end-to-end pipelines. A key component in classification is the use of a loss layer, which regulates the back-propagation error, for estimating prediction probabilities for each sample. For example, in CNNs the softmax loss function is typically used to minimise the difference between the predicted class probabilities and the ground-truth. Some models simultaneously predict both discrete emotions and continuous affect dimensions, such as arousal, valence, and strength of emotion \cite{toisoul2021estimation} (see Fig.\ref{emofan}). This aims to minimise data mislabelling and improve overall prediction accuracy.

 \subsection{Facial Expression Recognition}
FER systems begin with facial feature detection, whereby the face is identified and isolated. Methods such as the Viola-Jones algorithm, Histogram of Oriented Gradients (HOG), and Convolutional Neural Networks (CNNs) are used. Facial landmark detection identifies key points on the face, then feature extraction focuses on geometric features and appearance features. Traditional machine learning algorithms and deep learning models, especially CNNs, classify these features into emotional categories. CNNs are effective as they automatically learn and extract hierarchical features from raw pixel data \cite{LeCun2015}. The following section will discuss state-of-the-art research in FER with an emphasis on novelty, recurring themes, strengths, and limitations of current research. 

FER systems are classified into two categories: static image and dynamic sequence. While static methods encode spatial information from individual images, dynamic techniques use temporal relationships across frames within sequences \cite{Adyapady2023}. Historically, FER heavily relied on handcrafted features or shallow learning techniques such as Decision Trees \cite{Littlewort2011}, K-Nearest Neighbors (K-NN) \cite{Zhang2011}, and Support Vector Machines (SVM) \cite{Bartlett2006}. However, with the rise in emotion detection competitions such as FER2013 \cite{goodfellow2013challenges}, EMOCA \cite{Danecek2022}, and ABAW 2023 \cite{kollias_2023} a shift towards the use of deep learning techniques occurred. This has coincided with improvements in processing capabilities and network architectures, enabling the widespread adoption of deep learning methodologies. 

Models using pretrained Contrastive Language-Image Pretrained (CLIP) \cite{Radford2021clip} achieve remarkable results in FER. Using the joint embedding space of text and images, CLIP models can understand contextual information across modalities. By training on large datasets containing images paired with descriptions of emotions, CLIP learns to associate visual patterns with their emotional description. One such model which uses CLIP is DFER-CLIP \cite{zhao2023prompting}. This method combines both modalities, using a temporal model atop the CLIP image encoder. Temporal facial features are captured while using descriptions of facial behaviour instead of class names for the text encoder. It uses learnable prompts as context for descriptors of each facial expression class, enabling automatic learning of relevant context information during training. The model's pipeline involves extracting features from facial images or frames, and predicting facial expression descriptions. Furthermore, DFER-CLIP automates the generation of textual descriptors by prompting a language model with queries about useful visual features for each expression, culminating in comprehensive descriptions for classification. 

Attention is a key topic in FER with approaches such as self-attention, patch attention, and cross attention being utilised. EmoFan (see Fig.\ref{emofan}) uses attention mechanisms on facial landmarks and facial heat maps and achieves SOTA results. \cite{Liu_2023} uses patch attention and a pretrained ResNet-18 to extract the facial feature maps to overcome issues caused by occlusion for improved performance. \cite{Mao_2023} uses a similar approach by making use of window-based cross-attention mechanisms in conjunction with landmark detection, and multi-scale feature extraction. In comparison, \cite{gong2024enhanced} uses self-attention and a transformer to identify facial expressions in images or videos where the face is difficult to see. \cite{Danecek2022} addresses a shortfall in labelled datasets by incorporating an emotion recognition model into the 3D face reconstruction framework DECA\cite{tewari2020} This enables improved emotion reconstruction and classification, along with the use of their Emotion Consistency Loss. 

%The use of pretrained facial alignment models for facial landmark detection \cite{toisoul2021estimation}, or the use of 3D modelling to help with reenacting facial expressions \cite{danecek_2022} are two such approaches. Liu et al (2023) on the other hand utilized a modified pretrained ResNet-18 for the backbone of their model \cite{Liu_2023}. Poster++ \cite{Mao_2023} modified the architecture of Poster \cite{Zheng2023}, to address inter-class similarity and intra-class discrepancy in FER, and making a faster and simpler model. 

\subsection{Speech Emotion Recognition}

   \begin{figure}[hbt!] 

     \centering 

     \includegraphics[width=\textwidth]{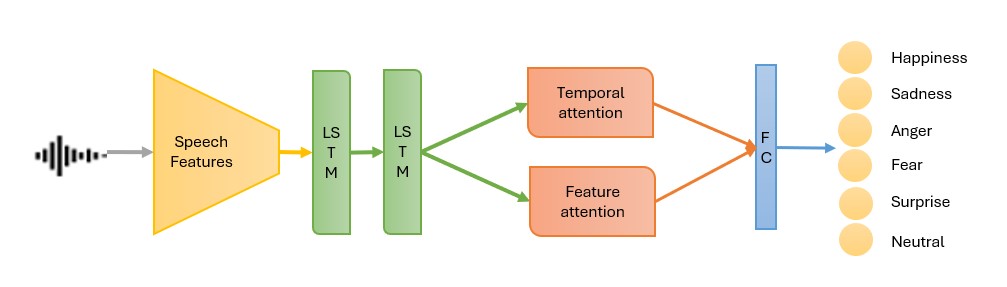} 

     \caption{The SER model processes frame-level speech features as input, using a 2-layer LSTM to generate outputs aligned with each frame's corresponding time. The LSTM's internal forget gate has been replaced by an attention gate. To differentiate emotional nuances across time and feature dimensions, the model applies a weighting operation separately on the LSTM's output along both the time and feature dimensions. These two weighted outputs are then fed into fully connected layers, and the final output from the softmax layer provides the classification result.\cite{Xie2023}} 

     \label{xie} 

 \end{figure} 
Recognising emotions in speech involves a multidisciplinary approach, integrating linguistics, psychology, and computer science \cite{Schuller2003}. Acoustic feature analysis, focusing on prosody and voice quality, plays a key role. Prosodic features, such as pitch, intensity, and speech rate, effectively indicate emotions. For example, happiness or excitement use higher pitch and greater variability, while sad voices use lower pitch and slower speech. Voice quality, including elements such as breathiness and tension, can also signal different emotions. Word choices and sentence structures, provide additional clues. Short, abrupt sentences can indicate anger, while longer, complex sentences might suggest calmness. Contextual analysis, considering the situational context and dialog history, is vital, as the same utterance can convey different emotions depending on the context \cite{Scherer2003}. 

Transformer based model ESCM \cite{Yang2023}, achieved state-of-the-art results in SER by adjusting emotions and semantics based on context. They achieve this by using Graph Convolutional Network (GCN) to find correlations between words in spoken coversations. In contrast, \cite{Xie2023} (see Fig.\ref{xie}) introduces a novel approach to speech emotion recognition by integrating attention mechanisms into Long Short Term Memory (LSTM) models. By prioritising relevant information across both time and feature dimensions, the attention-based LSTM architecture improves performance in SER. The use of frame-level features provide a comprehensive representation of emotional content, contributing to the model's accuracy. \cite{gong2024enhanced} use Large Language Models (LLMs) and weakly-supervised learning to label the emotions in speech data, which contributes to the effectiveness of their SER model.

Further innovations in time-frequency analysis have also improved SER. For instance, the fast Continuous Wavelet Transform (fCWT) enables high-resolution analysis of non-stationary speech signals, balancing temporal and spectral features. When combined with Deep Convolutional Neural Networks (DCNNs), this approach enhances the extraction of paralinguistic information, offering robust real-time performance while overcoming limitations of traditional methods like the Short-Term Fourier Transform (STFT) \cite{van2023speech}.

\subsection{Text Sentiment Recognition}
  \begin{figure}[hbt!] 

     \centering 

     \includegraphics[width=\textwidth]{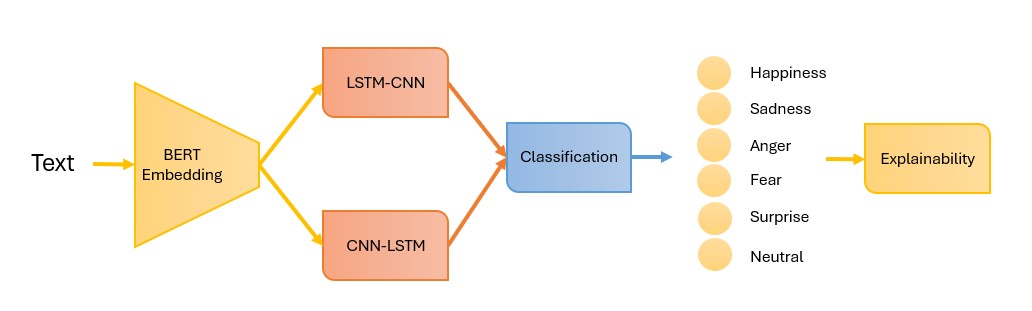} 

     \caption{The TER system by\cite{Kumar2022} uses a BERT-based dual-channel pipeline for text emotion recognition. First, input sentences are converted into contextual embeddings with a pre-trained BERT model. These embeddings are then processed through two parallel channels: one uses CNN for feature extraction followed by BiLSTM for capturing sequence information, while the other uses BiLSTM first, followed by CNN. The outputs from both channels are concatenated and passed through dense layers for emotion classification. An explainability module further interprets the model's predictions by analysing emotion embedding clusters.} 

     \label{kumar} 

 \end{figure} 
TER focuses on the identification and classification of emotions expressed in textual data using Natural Language Processing models (NLP). NLP models enable machines to understand, interpret, and generate text \cite{Jurafsky2009}. Bidirectional Encoder Representations from Transformers (BERT) \cite{Devlin2019} are used in most modern NLP models \cite{Vaswani2017}. These models are useful for TER due to their ability to capture contextual data and decipher emotions in text, enabling SOTA performance. Campagnano et al. \cite{Campagnano2022semantic} combines BERT encodings with bidirectional LSTM layers to achieve robust emotion classification, particularly in semantic role labelling tasks. \cite{Koptyra2023} use a modified BERT-based architecture to classify emotions for individual sentences and entire texts. \cite{Bianchi2022} use a BERT model trained on data from 100 languages as well as X (formerly Twitter), to detect emotions on social media platforms. In contrast, \cite{Kumar2022} (see Fig.\ref{kumar}) use LSTM and a CNN based model for TER. The use of CNN-LSTM channels extracts both local and global contextual information from input text, working for diverse text inputs. \cite{Koptyra2023, Bianchi2022} address multilingual emotion recognition, developing models and datasets capable of working across languages. As seen in this analysis there is a distinct lack of recent research into TER, highlighting the need for updated studies to address current challenges and advancements in the field.

\section{Emotion Generation for Faces, Speech, and Text}

This section will discuss generated content for faces - which will focus on animated face generation, speech - taking the nuances of audio from one speaker and converting to another voice, and text - the generation of realistic text. Emotion recognition models are sometimes used for training \cite{Li2020dual}, and evaluating \cite{Schuller2004} these models to generate accurate emotional content. Emotion recognition datasets are also utilised for emotion generation models \cite{Zeng2020}. A recent challenge with creating emotionally realistic generated content comes from negativity in public's perception due to media hype surrounding stealing of identities \cite{Matton2019}, deepfakes \cite{Dolhansky2020}, and the rapid rate in which models are being released \cite{Radford2021clip}. This consideration has the capacity to hinder research in these fields due to restrictions on the availability of models for researchers \cite{Goodfellow2019}, due to the fear they will fall into the wrong hands. This section will discuss SOTA methods for these modalities, and will discuss the strengths and limitations of current research.

\subsection{Facial Expression Generation}
%overall
 \begin{figure}[hbt!] 

     \centering 

     \includegraphics[width=\textwidth]{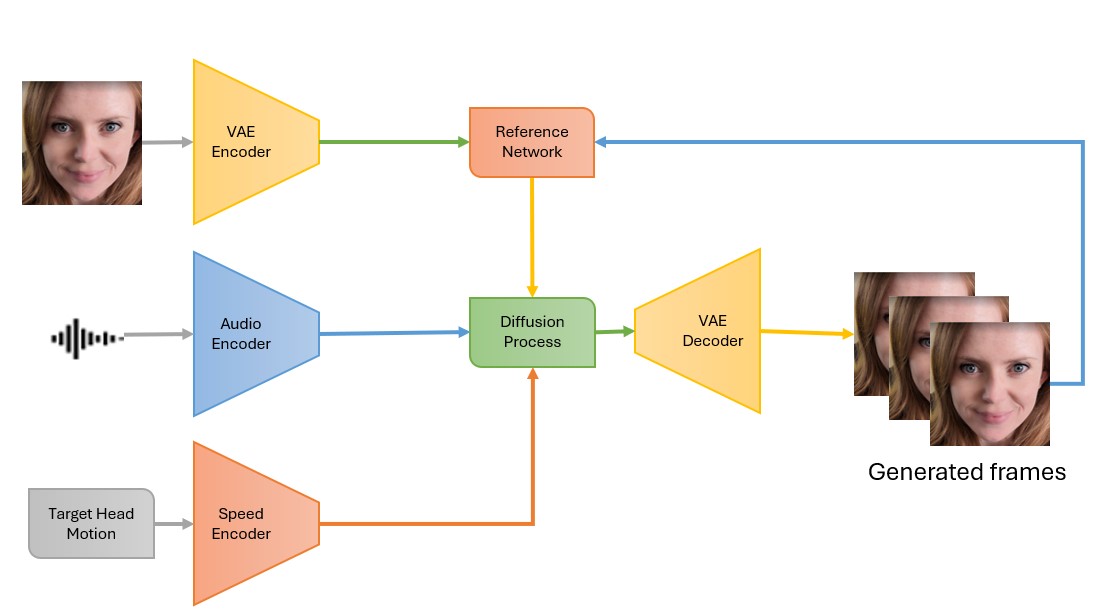} 

     \caption{
 In the place of 3D modelling, EMO utilising Stable Diffusion for generating new frames. The  pipeline consists of a Backbone Network paired with a ReferenceNet to maintain identity consistency, audio-attention layers to synchronise facial expressions with audio tonalities, and temporal modules to ensure smooth transitions across frames. Weak control signals, such as a Face Locator and Speed Layers, provide loose guidance for facial positioning and movement velocity, achieving natural and stable head motions across clips.\cite{Tian2024emo}} 

     \label{Tian} 

 \end{figure}

FEG and face manipulation techniques have been around for years, present on mobile phone apps such as Instagram \cite{instagram}, SnapChat \cite{snapchat}, and AI photo editors such as FaceApp \cite{faceapp} and others. The release of visually appealing talking-head models such as VASA \cite{xu2024vasa} and EMO-Live \cite{Tian2024emo}, have further bolstered public interest in this research area. Talking-head animation refers to models which take as input an image of a person, and generates new frames using audio \cite{Tian2024emo} (see Fig.\ref{Tian}), video \cite{paraperas2021emo}, or text \cite{li_2020} to guide the facial expressions. The manipulation of facial expressions through prompts is another new area of research \cite{Gan2023,Chun2021egsdfa,chen2023expressive}. FEG models often focus on prioritising the manipulation of the mouth, eyes, or poses \cite{zhou2021pose,prajwal2020lip,wang2023seeing,Park2022}, while others focus on overall realism \cite{Tian2024emo,xu2024vasa, zhang2023sadtalker}. With mouth movements now achieving realism pretrained SOTA models such as Wav2Lip \cite{prajwal2020lip} are incorporated into larger models to guide the lip movements, while the model focuses on poses and facial expressions \cite{zhang2023sadtalker}. \cite{Tian2024emo,paraperas2021emo} use 3D face modelling techniques and reconstruction methods to capture detailed facial geometry. This allows for accurate expression synthesis and emotion manipulation. Similarly, other methods use 3D registration and mesh-based representations to achieve realistic Face Expression generation \cite{Niinuma2022,Li2021starganv2}.

Generative Adversarial Networks (GAN) are used for generating animations \cite{Gan2023,Vougioukas2020} due to their ability to create realistic synthetic content. GANs are trained by generating content through a generator network, then using a discriminator network to predict if the generated content is real or not. For Face Expression generation, GANs are combined with other models to generate realistic facial expressions in talking-head animation generation \cite{Kong2021dualpath,Ling2020,Yan2023gan,Chun2021egsdfa,Niinuma2022}. For example, \cite{Chun2021egsdfa} employ LSTM networks and a GAN for speech-driven animation. \cite{Kong2021dualpath} use a GAN to guide the generation process of emotional animations, and preserve the identity of the target face. \cite{Ling2020} focuses on facial expression manipulation using a modified U-Net structure with GANs and achieves precise emotion manipulation. \cite{Wang2021} use GANs and attention mechanisms as the backbone of their text to talking-head generation framework. Meanwhile, \cite{Pikoulis2023} and \cite{Gan2023} utilise GANs in their methodologies for efficient emotional manipulation. Additionally, \cite{Niinuma2022} use a GAN for personalised facial expression manipulation. 
\cite{Tian2024emo} use Diffusion models for generative power and extensive control over the generation of animations. Diffusion models iteratively refine a noisy image into a high-quality sample. This refinement allows for the generation of highly realistic facial expressions, while maintaining control over intensity, duration, and subtle movements. By conditioning the diffusion process on desired expression labels or latent codes, these models produce specific facial expressions with remarkable realism. As diffusion models capture uncertainty during generation, this enables the synthesis of realistic variations. \\\\
Attention's ability to focus on important facial regions and generate realistic facial expressions has enabled them to become a key part of face generation architectures. In \cite{chen2023expressive}, attention mechanisms ensure the generated facial animations accurately capture the speaker's gestures and facial expressions. \cite{Tian2024emo} (see Fig.\ref{Tian}) integrates attention mechanisms into the pipeline to improve the quality and synchronisation of talking portrait videos, attention mechanisms are utilised to refine motion dynamics and speed adjustments. This method achieves realistic talking portrait videos which closely align with the input audio content. \cite{Wang2021} use attention gate and self-attention mechanisms in their text-based talking-head generation framework. By incorporating these mechanisms their model manipulates Action Unit-related embeddings, leading for accurate and expressive facial animations synchronised with input text.\\
CLIP with its multimodal capabilities is useful for facial animation generation tasks. By inputting textual prompts to describing desired emotional states along with images associated with those emotions, CLIP can generate images reflecting the specified emotions. This allows the model to learn associations between text and images which improves its ability to generate content with realistic emotions. TalkCLIP by\cite{ma2023dreamtalk} generates realistic talking head videos of a target speaker with specific speaking styles. Their model utilises CLIP embeddings and an adaptor network to map text descriptions, to speaking style codes. \\

Furthermore, researchers have explored the ability to control the generation of emotions on the faces through various inputs such as speech, video, facial reenactment, and text. Speech data is the most common input medium whereby an animated face video is generated using the emotions in the speech \cite{Chun2021egsdfa,Gan2023,Tian2024emo} (see Fig.\ref{Tian}). Video is used as an input in architectures where the face is changed to a target face using facial reenactment methods \cite{paraperas2021emo}, or the emotions are manipulated via facial reenactment from a static image \cite{Kong2021dualpath}. However, the synchronisation of speech and facial animations rely on robust phoneme processing within the architectures \cite{Wang2021}. Using text as a input is a relatively unexplored field which enables the generation  of Face Expression generation based on the emotion content of textual dialogue \cite{Li2021starganv2}. Other researchers have explored methods to directly control the emotions on the output videos using CLIP text prompts \cite{Gan2023,ma2023dreamtalk}.

\subsection{Speech Emotion Generation}
 \begin{figure}
    \centering
    \includegraphics[width=\linewidth]{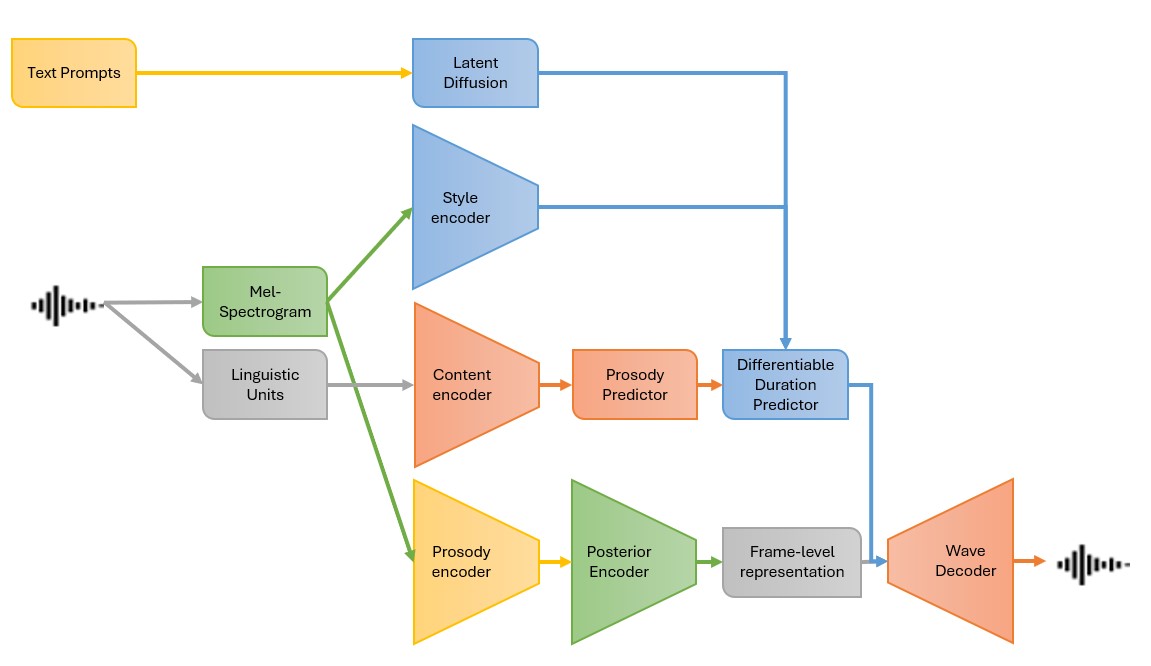}
    \caption{The PromptVC pipeline uses a latent diffusion model for voice style conversion using natural language prompts. During training, a style encoder extracts a global style vector from the input mel-spectrogram, while HuBERT-based discrete tokens capture linguistic content, refined by a differentiable duration predictor for accurate timing. A prosody encoder models phoneme-level prosody to enhance expressiveness. The latent diffusion model, conditioned on text embeddings, generates the style vector from noise, enabling flexible and precise style control.\cite{yao2024promptvc}}
    \label{promptvc}
\end{figure}
One element of SEG, known as voice conversion, speech-to-speech synthesis, or speech reenactment, involves the transformation of speech signals to modify the vocal characteristics of one speaker to resemble another or to produce entirely synthetic voices. These methods form the basis of SEG, whereby the emotions in a target voice can be changed through prompts \cite{yao2024promptvc}(see Fig.\ref{promptvc}), or by the emotions in a target voice through using a emotional reference voice \cite{Park2023}.

Recent advancements in AI have led to the development of synthetic voices that are almost indistinguishable from human speech. Achieving realism in generated speech involves capturing natural intonation, rhythm, and emotion. Advanced systems, such as those by ElevenLabs \cite{elevenlabs}, use SOTA deep learning techniques to produce high-quality, realistic speech. These systems generate voices that sound authentic and carry unique characteristics associated with individual speakers. This section reviews recent advancements in SEG methodologies.

SEG models use phonetic content, including emotional cues, from a source voice to synthesize audio in a target voice while retaining desired stylistic characteristics \cite{Park2023}. A common approach in SEG involves using language models like BERT \cite{Devlin2019} for extracting contextualized representations of linguistic content, thereby enabling precise alignment between source and target voices. BERT embeddings contribute to the controllability and realism of SEG systems, facilitating accurate transformations in speech style and characteristics, which allows synthetic speech to be tailored to specific emotions \cite{Xu2023, Park2023, Lin2023, Maimon2022}. Traditional approaches often rely on text-based conditioning using transcripts \cite{taylor2009text}; however, recent methods, such as that by \cite{Maimon2022}, employ discrete representations for phonetic content. This enables the capture of non-textual cues, such as laughter, and supports diverse linguistic applications. Additionally, \cite{Lin2023} propose an architecture that integrates source and target encoders with a decoder, preserving critical linguistic and speaker features throughout the conversion process to ensure the synthesized speech remains natural and true to the source.\\\
SEG also benefits from adversarial training techniques inspired by GANs \cite{yang2021ganspeech}. In these frameworks, a discriminator differentiates between target voice samples and synthesized speech, prompting the model to generate speech that convincingly reconstructs the source content while mimicking the target speaker's characteristics.The DDDM-VC model \cite{choi2024dddm} introduces a novel approach for SEG, enhancing controllability by decoupling and independently processing attributes such as content, pitch, and timbre. Through attribute-specific denoising, DDDM-VC achieves high-precision voice style transformations, while the inclusion of prior mixup techniques strengthens robustness in voice adaptation, especially in zero-shot scenarios. This disentangled structure enables DDDM-VC to maintain speaker fidelity and naturalness in synthesized voices across a variety of speaker styles . Similarly, PromptVC by \cite{yao2024promptvc} (see Fig.\ref{promptvc}), uses a latent diffusion model for voice style conversion using natural language prompts. This enables precise control over the attributes in the generated speech. Another method uses Contrastive Predictive Coding (CPC) features to enhance the quality of synthesised speech \cite{Park2023}, which is a self-supervised learning technique for predicting future utterances in latent space. Similarly, \cite{Bous2021} preserves time-synchronisation and fundamental frequency information to maintain the naturalness of converted speech. Finally, two-stage training schemes are frequently used to align hidden representations between source and target speech. The initial stage focuses on reconstructing single utterances to establish alignment, followed by a second stage where multiple utterances refine the conversion process \cite{Lin2021}. This progressive refinement enhances the model's adaptability, improving performance in scenarios with significant divergence between source and target speech characteristics. 

\subsection{Text Sentiment Generation} \label{EmotionalText}% mention social media such as blogs and tweets are used, BERT based models, ChatGPT
\begin{figure}
    \centering
    \includegraphics[width=0.90\linewidth]{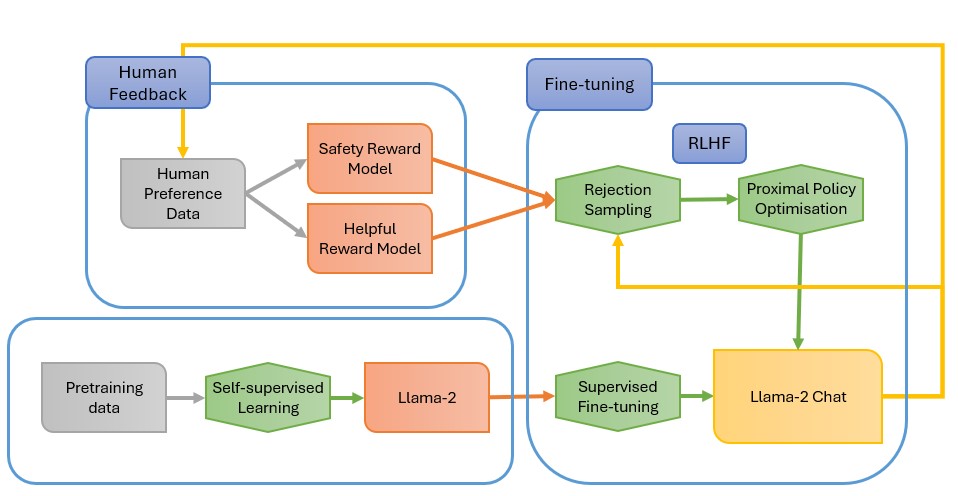}
    \caption{The LLaMA pipeline involves pre-training transformer-based models on large textual datasets, followed by task-specific fine-tuning through supervised learning, and reinforcement learning with human feedback (RLHF). Efficient fine-tuning is achieved using methods such as QLoRA, which significantly reduce computational requirements. The model is iteratively optimised and evaluated to attain state-of-the-art performance across various applications. \cite{llama}}
    \label{llama_chat}
\end{figure}

TSG models work from a user interface by taking input text, and generating a response (see Fig. \ref{llama_chat}). TSG have the ability to alter the emotional content in existing text. Large Language Models (LLMs) such as ChatGPT \cite{achiam2023gpt}, Llama \cite{llama} (see Fig.\ref{llama_chat}), Gemini \cite{reid2024gemini} can create text with emotions and personality which can pass for human writing. Ensuring accurate grammar and syntax, a diverse and contextually appropriate vocabulary, and consistency in style, tone, and information are all important for TSG. Additionally, typographical errors, realistic mistakes, smooth transitions between ideas and a deep understanding of context also contribute to the text's realism. 

%Advanced natural language generation techniques and feedback loops for refinement further improve this. Field-specific knowledge and contextual understanding are also essential for generating text that aligns with the conversation or topic at hand.
Until recently Recurrent Neural Networks (RNNs) \cite{rumelhart1986learning} were used extensively in text generation due to their ability to handle sequential data by maintaining an internal memory. However, traditional RNNs suffer from the vanishing gradient problem, which impedes long-range dependencies. They also struggled to work on long sentences \cite{hochreiter1997long}. Researchers attempted to combat this by running the RNNs both forward and backward over the textual data \cite{schuster1997bidirectional}, which did not rectify the problem. These limitations led to the development of Long Short-Term Memory (LSTM) networks, a variant of RNNs. LSTMs employ architectures with gated mechanisms, including input, output, and forget gates, enabling them to learn and retain long-term dependencies in sequential data \cite{hochreiter1997long}. This feature makes LSTMs particularly ideal for tasks requiring memory over extended sequences, such as text generation. Another architecture used for text generation are Sequence-to-Sequence (Seq2Seq) models \cite{sutskever2014sequence}, which consist of an encoder and a decoder. Seq2Seq models have shown proficiency in generating coherent and contextually relevant text, making them valuable for emotional text generation tasks. Generative Adversarial Networks (GANs) \cite{goodfellow2014generative}, used mostly in computer vision, have also emerged as useful for text generation tasks. The generator produces synthetic text data, while the discriminator evaluates the authenticity of the generated text. Used in conjunction with the above algorithms, attention mechanisms enable models to focus on relevant parts of the input text sequence when generating a response. Attention mechanisms allow models to weigh the importance of each word in the input sequence dynamically as they generate each word in the output sequence \cite{bahdanau2014neural}. For example, in the Seq2Seq model, attention mechanisms help align the encoder hidden states with the decoder hidden states at each time point, ensuring the model attends to the most relevant parts of the input sequence when generating each word in the output sequence \cite{bahdanau2014neural}.

%However, despite these advancements, emotional text generation remains a challenging field within NLP. Emotions are nuanced and subjective, posing difficulties for humans and machines alike. Often, these systems rely on surface-level features such as punctuation or word choice to infer emotional states, which may lead to imprecise results. Moreover, human interpretation of emotional text varies widely based on individual experiences, cultural background, and personal writing style, adding complexity to the task. This means creating datasets with accurate human annotations is challenging. 

To address these challenges researchers are exploring various approaches. One approach involves fine-tuning pretrained language models such as ChatGPT \cite{achiam2023gpt} for emotion-specific tasks \cite{Goswamy2020human}. This approach uses datasets annotated with emotional labels to train the model to associate linguistic patterns with emotional states. During fine-tuning, adjustments are made to the model's parameters through additional training iterations on emotional text datasets. Developing models with an understanding of contextual cues is essential for accurate emotional text generation. This involves considering factors such as the broader narrative, speaker intent, and audience context to generate realistic text. 

\subsection{Generative Models with Emotion Control}
This section will examine methodologies for implementing emotion control within FEG, SEG, and TSG. Emotion control, in this context, pertains to the systematic generation of content—spanning animations, speech, and textual outputs—characterised by realistic and contextually appropriate emotional expressions. These emotions are elicited or guided through specific prompts or control mechanisms, ensuring that the generated outputs align with intended affective states. The discussion will encompass techniques used to encode, manipulate, and render emotions, as well as the underlying computational models that enable nuanced emotional dynamics across various modalities.

%\begin{figure}[hbt!] 

  %  \centering 

    %\includegraphics[width=0.75\textwidth]{tian_fig.PNG} 

   % \caption{ The EMO-Live method by \cite{Tian2024emo} comprises two main stages: Frames Encoding and Diffusion Process. In Frames Encoding, ReferenceNet extracts features from the reference image and motion frames. In the Diffusion Process, a pretrained audio encoder processes audio embedding. A facial region mask guides facial imagery generation with multi-frame noise. The Backbone Network facilitates denoising, incorporating Reference-Attention and Audio-Attention mechanisms to maintain identity and modulate movements. Temporal Modules adjust motion velocity.} 

    %\label{Tian} 

%\end{figure} 

\subsubsection{Audio Driven Face Expression Generation} 

Fig. \ref{Tian} shows audio driven Face Expression generation by \cite{Tian2024emo}. This method for Face Expression generation takes a reference image as input which is put through a frames encoder. Next, a feature extraction network, called ReferenceNet extracts detailed features from the reference image and after the first iteration, the motion frames, to preserve the identity from the reference image. The architecture then progresses to the diffusion stage where a pretrained audio encoder processes the input voice audio clip, extracting voice features which influence the facial movements and expressions. The Backbone Network, using reference-attention and audio-attention mechanisms, denoises the input data and generating realistic video frames. This comprehensive network architecture ensures the generated video frames sync with the provided audio content. Speed layers fine-tune temporal modules and control head motion across clips, improving consistency and stability in the generated videos. 
%The user interface is initiated with a reference image and an audio clip containing the desired voice content. Users can next configure model settings according to their specific requirements and preferences. This step involves adjusting parameters such as the desired length or duration of the generated video, the level of detail in facial expressions, and the synchronisation between audio and video elements. By customising these settings, users can tailor the generated videos to suit their intended artistic or communicative objectives. This iterative approach allows users to experiment with different configurations and refinements until they achieve a satisfactory outcome that meets their expectations and requirements.

%\begin{figure}[hbt!] 

  %  \centering 

    %\includegraphics[width=0.5\textwidth]{Li_fig.PNG} 

  %  \includegraphics[width=0.5\textwidth]{Li_fig2.PNG} 
    
 %   \caption{This diagram illustrates the pipeline for generating photo-realistic talking-head videos from text inputs, a method by \cite{Li2021starganv2}. Beginning with time-aligned text, specialised modules translate it into animation parameters for mouth, upper face, and head pose. Using a motion capture dataset, the system ensures realistic as facial expressions and head movements. Neural networks are trained to optimise parameter generation. A style-preserving landmark generator aligns facial features with speaker-specific characteristics. Finally, video synthesis renders textures for a realistic talking-head video output, embodying the nuances of the input text.} 

  %  \label{li} 

%\end{figure} 

\subsubsection{Text Driven Face Expression Generation}
The text-based talking-head generation framework by \cite{Li2021starganv2} uses neural networks tailored to different aspects of generating Face Expression animations from textual inputs. Gmou, dedicated to animating mouth movements from phonemes, uses a structure based on CNNs for efficient parallel computation and is trained using a combination of L1 loss and Least Squares Generative Adversarial Network (LSGAN) loss. Similarly, Gupp and Ghed utilise encoder-decoder network structures to synthesize upper face parameters and head pose, respectively, from input words, training with analogous loss functions to ensure realistic outputs. The Style-Preserving Landmark Generator, Gldmk, uses a multi-linear 3D Morphable Model (3DMM) and a fully-connected network to ensure consistency and accuracy in facial expressions, incorporating a unique mapping technique to preserve speaker-specific styles. %Together, these networks work within the framework to generate cohesive and expressive talking-head videos, seamlessly inTSGrating textual inputs with realistic facial animations.

%Users provide time-aligned text inputs of dialogue or speech for the generated talking-head video. The system then processes these inputs through its pipeline, using the trained neural networks to generate animation parameters and synthesise the final video output. Users can then seamlessly view and analyse the generated videos, which accurately depict the subtleties of the speaker's facial expressions and head movements as conveyed by the input text.

%\begin{figure}[hbt!] 

  %  \centering 

    %\includegraphics[width=\textwidth]{para_fig.PNG} 
    
    %\caption{\cite{paraperas2021emoemo}} 

  %  \label{para} 

%\end{figure} 

\subsubsection{Video Driven Face Expression Generation}
%The method uses a sophisticated framework called the Neural Emotion Director (NED) to 

NED by \cite{paraperas2021emo} allows manipulation of Face Expressions in in-the-wild videos while preserving natural speech-related mouth motion. The Face Analysis module incorporates preprocessing steps such as face landmark detection, segmentation, and resizing, alongside 3D Morphable Models (3DMMs) for accurate estimation of 3D face geometry. The Expressions Translator, a GAN, utilises a recurrent network with LSTM units to convert sequences of facial expressions into desired emotions, while maintaining the original mouth motion. A encoder extracts emotion-related style vectors from the input sequences, while the Mapping Network generates style vectors associated with target emotions. A neural face renderer generates realistic frames, incorporating techniques such as multi-band blending for seamless integration of generated faces into the original backgrounds. This ensures the manipulated facial expressions seamlessly blend into real-world scenarios. During testing, N-length sliding windows are applied frame by frame, with the sequences processed through the Expressions Translator. The conditional style vector is either generated by the Mapping Network or extracted from a reference video, allowing for flexible manipulation of emotions in facial videos.

%\begin{figure}[hbt!] 
  %  \centering 

    %\includegraphics[width=0.5\textwidth]{Gan_fig.PNG} 
    %\includegraphics[width=0.5\textwidth]{Gan_fig2.PNG} 
    
  %  \caption{\cite{Gan2023}} 

 %   \label{ganfig} 

%\end{figure} 

\subsubsection{Emotion Prompted Face Expression Generation}
EAT by \cite{Gan2023} takes in an image of a target face, speech, and an emotion prompt such as happy, sad, or angry, to generate animated videos. The model first trains the CLIP model on emotion labelled datasets to learn audio-visual correlations. This pre-training phase uses enhanced latent representations and a transformer model. Enhanced latent representations capture intricate facial expressions, incorporating identity-specific canonical keypoints, rotation, translation, and expression deformation components. The transformer model predicts synchronised expression deformations from audio inputs and predicts head pose features, and latent source image representations. Next, three primary modules—Deep Emotional Prompts, Emotional Deformation Network (EDN), and Emotional Adaptation Module (EAM)—play integral roles in the emotional adaptation. Deep Emotional Prompts inject emotion-guided expression generation into the model, using latent codes sampled from a Gaussian distribution to provide crucial emotional guidance. EDN complements this by predicting emotion-related expression deformations. EAM further refines the visual quality of generated videos by generating emotion-conditioned features. The architecture also accommodates zero-shot expression editing, which allows text-guided manipulation of talking-head videos without the need for extensive emotional training data. Using the CLIP model, the system aligns generated expressions with textual descriptions, offering users control over the emotional content of the videos. 

\subsubsection{Speech Emotion Generation Model}
The architecture in \cite{Park2023} comprises three main components: source encoder, target encoder, and a decoder. The source encoder uses Wav2Vec 2.0 \cite{baevski2020wav2vec}, a pretrained feature extractor, to capture speech representations from the source utterance. The target encoder processes log mel-spectrograms of utterances from the target speaker, and the decoder consists of transformer layers using both self-attention and cross-attention. A linear projection layer contributes to the final prediction of the desired output voice, following a non-autoregressive approach. The model is trained using a two-stage approach. In the first stage, single utterances from both the source and target speakers are used to reconstruct the log mel-spectrogram of the utterance. In the second stage, multiple utterances, typically 10, from the target speaker are concatenated and fed into the target encoder. Simultaneously, a single utterance from the source speaker is fed into the source encoder.

\subsubsection{Text Sentiment Generation Model}
A model \cite{Goswamy2020human} built upon ChatGPT2 \cite{radford2019language}, has been trained to generate text with specific emotions. The ChatGPT2 model is fine-tuned with text samples annotated with affective labels or sentiment scores. The Plug and Play Language Model (PPLM) framework is integrated into the ChatGPT2 architecture to enable attribute-controlled text generation. PPLM incorporates perturbation and optimisation mechanisms during training, enhancing the model's ability to generate text with specific affective attributes. The model's loss functions include terms which encourage the generation of text with desired emotional attributes and intensity levels. Users specify the desired emotional tone or topic, and the intensity of the emotion desired. The model uses specified attributes and intensity levels to control the content and tone during the text generation process.

\subsubsection{Text Sentiment Generation Chatbot}
The Empathetic Semantic Correlation Model (ESCM) by \cite{Yang2023} generates empathetic responses in dialogues by understanding emotions and semantics. It includes three components: a context encoder, a dynamic correlation encoding module, and an emotion and response predicting module. The dynamic correlation encoding module features dynamic emotion-semantic vectors and a correlation Graph Convolutional Network, adjusting emotions and semantics based on contextual cues. The emotion and response predicting module uses context semantics and correlations to predict emotions and generate empathetic responses. During training, ESCM optimises parameters using multiple loss functions and supervised learning on annotated datasets. In use, ESCM processes dialogue context, adjusts to contextual cues, and continuously learns to provide accurate, empathetic responses.

\section{Evaluation}

This section provides an overview of the metrics used to evaluate ER and EG models across facial, speech, and textual modalities. It explores various evaluation techniques to determine their effectiveness in measuring model performance and accuracy. Furthermore, the comparative analysis within this section examines state-of-the-art methods to identify the most effective approaches. By synthesising findings from recent studies, this evaluation aims to uncover the strengths and limitations of current evaluation frameworks, thereby highlighting which models are most proficient at recognising and generating emotional expressions across different modalities.

\subsection{Evaluation Metrics}

Evaluation metrics are essential for assessing the performance of emotion recognition and generation models across different modalities. This section highlights the most widely used metrics in facial, speech, and text, emotion recognition and generation.

\subsubsection{Common Metrics}

\begin{itemize}
    \item \textbf{Accuracy}: This metric measures the proportion of correctly classified instances among the total instances. It provides a basic overview of model performance but does not account for class imbalances, which can lead to misleading results.
    \begin{equation}
    \text{Accuracy} = \frac{\text{Number of Correct Predictions}}{\text{Total Number of Predictions}}
    \end{equation}

    \item \textbf{F1 Score}: The harmonic mean of precision and recall, providing a balanced measure of a classifier’s performance, particularly in cases with imbalanced datasets. The F1 score is crucial for understanding the trade-off between precision and recall.
    \begin{equation}
    F1 = \frac{2 \times \text{Precision} \times \text{Recall}}{\text{Precision} + \text{Recall}}
    \end{equation}

    \item \textbf{Precision}: Measures the proportion of true positive predictions out of all positive predictions, indicating the accuracy of positive predictions in identifying emotional expressions.
    \begin{equation}
    \text{Precision} = \frac{\text{True Positive}}{\text{True Positive} + \text{False Positive}}
    \end{equation}

    \item \textbf{Recall}: Measures the proportion of true positive predictions out of all actual positives, reflecting the model's ability to identify relevant instances. High recall is essential in applications where missing a positive instance can have significant consequences.
    \begin{equation}
    \text{Recall} = \frac{\text{True Positive}}{\text{True Positive} + \text{False Negative}}
    \end{equation}

    \item \textbf{Mean Opinion Score (MOS)}: Often used in evaluating generated speech and facial expressions, this metric assesses perceived quality by averaging ratings given by human evaluators on a numerical scale, providing a subjective measure of output quality.\\
\end{itemize}
%     \item \textbf{Confusion Matrix}: This tool provides a detailed breakdown of the performance of a classification model, showing the numbers of true positives, true negatives, false positives, and false negatives, allowing for a comprehensive evaluation of model performance.\\
% \end{itemize}

\subsubsection{Metrics for face systems}
\begin{itemize}
\item \textbf{Structural Similarity Index (SSIM}, \ref{Eq:ssim}, is used to assess the similarity between two images. It takes into account luminance, contrast, and structure of the images. SSIM is defined as:
\begin{equation}
    \text{SSIM}(x, y) = \frac{{(2\mu_x\mu_y + C_1)(2\sigma_{xy} + C_2)}}{{(\mu_x^2 + \mu_y^2 + C_1)(\sigma_x^2 + \sigma_y^2 + C_2)}}
    \label{Eq:ssim}
\end{equation}
%Where $\mu_x$, $\mu_y$, $\sigma_x^2$, $\sigma_y^2$, and $\sigma_{xy}$ are the mean, variance, and covariance of the pixel intensities of images $x$ and $y$, respectively, and $C_1$ and $C_2$ are constants to stabilize the division with weak denominator.

\item \textbf{ Fréchet Inception Distance score (FID)}, \ref{Eq:fid}, evaluates the quality of generated images in generative adversarial networks (GANs). It measures the similarity between the distribution of real images and generated images in a feature space learned by a pretrained deep convolutional neural network. FID is defined as:
\begin{equation}
    \text{FID} = \|\mu_x - \mu_y\|^2 + \text{Tr}(\Sigma_x + \Sigma_y - 2(\Sigma_x\Sigma_y)^{1/2})
        \label{Eq:fid}
\end{equation}
%Where $\mu_x$, $\mu_y$ are the mean feature vectors and $\Sigma_x$, $\Sigma_y$ are the covariance matrices of the feature representations of real and generated images, respectively. Tr denotes the trace of a matrix, and $\|^2$ represents the squared Frobenius norm of a matrix.

  \item \textbf{Cumulative Probability Blur Detection (CPBD)} \\
    CPBD quantifies image blur by analysing edge sharpness and comparing edge gradient profiles to perceptual thresholds. A higher CPBD score indicates a clearer image with less blur.
    \[
    \text{CPBD} = \frac{1}{N} \sum_{i=1}^N \mathcal{P}(e_i)
    \]

    \item \textbf{Cosine Similarity (CSIM)} \\
    CSIM measures the similarity between two vectors, such as feature embeddings of source and generated faces. Values range from \(-1\) to \(1\), where \(1\) indicates identical direction and maximum similarity.
    \[
    \text{CSIM} = \frac{\mathbf{A} \cdot \mathbf{B}}{\|\mathbf{A}\| \|\mathbf{B}\|}
    \]

    \item \textbf{Mouth Landmark Distance (M-LMD)} \\
    M-LMD evaluates the average difference in lip keypoint positions between reference and generated videos. It reflects the overall accuracy of lip synchronisation in generated content.
    \[
    M\text{-LMD} = \frac{1}{T} \sum_{t=1}^T \frac{1}{K} \sum_{k=1}^K \| \mathbf{p}_{t,k}^{\text{ref}} - \mathbf{p}_{t,k}^{\text{gen}} \|
    \]

    \item \textbf{Face Landmark Distance (F-LMD)} \\
    F-LMD calculates the keypoint difference between reference and generated faces. It provides insights into face synchronisation.
    \[
    F\text{-LMD}(t) = \frac{1}{K} \sum_{k=1}^K \| \mathbf{p}_{t,k}^{\text{ref}} - \mathbf{p}_{t,k}^{\text{gen}} \|
    \]
\end{itemize}
 
\subsubsection{Metrics for speech and text systems}

\begin{itemize}
    \item \textbf{Word Error Rate (WER)}: Commonly used in speech and text systems, WER quantifies the rate of incorrect words generated by the system compared to a reference transcript. Lower WER scores indicate better system performance in speech and text generation tasks.
    \begin{equation}
    \text{WER} = \frac{\text{Number of Word Errors}}{\text{Total Number of Words in Reference Transcript}}
    \end{equation}

    \item \textbf{Character Error Rate (CER)}: Similar to WER, this metric measures the rate of incorrect characters generated in speech and text systems compared to the reference transcript. It provides a more fine-grained evaluation of textual accuracy, particularly useful in text-based emotion recognition systems.
    \begin{equation}
    \text{CER} = \frac{\text{Number of Character Errors}}{\text{Total Number of Characters in Reference Transcript}}
    \end{equation}

    \item \textbf{Equal Error Rate (EER)}, \ref{eq:eer}, is a point where the false acceptance rate (FAR) and false rejection rate (FRR) are equal in a speaker systems. It represents the operating point where the system's performance is balanced. 
\begin{equation}
    \text{EER} = \text{FAR} = \text{FRR}
    \label{eq:eer}
\end{equation}

 \item \textbf{Mel-cepstral distortion (MCD)}, \ref{eq:mcd}, quantifies the difference between two sets of mel-frequency cepstral coefficients (MFCCs) for speech tasks. 
\begin{equation}
    \text{MCD} = \frac{1}{N} \sum_{i=1}^{N} \| X_i - Y_i \|
    \label{eq:mcd}
\end{equation}

    \item \textbf{Perplexity}: A key metric in text generation, perplexity measures how well a language model predicts a sample of text. It reflects the average branching factor of the model, with lower perplexity indicating better performance.
    \begin{equation}
    \text{Perplexity} = 2^{-\frac{1}{N} \sum_{i=1}^{N} \log P(x_i)}
    \end{equation}

    \item \textbf{Sentiment Accuracy}: For text-based emotion recognition, sentiment accuracy measures how accurately a model classifies the overall emotional tone or sentiment of a text (e.g., positive, negative, neutral). This metric is widely used in applications such as sentiment analysis and Text Sentiment generation.\\
    
    \item \textbf{BLEU (Bilingual Evaluation Understudy Score)}: Commonly used in text generation systems, BLEU compares the generated text to a reference by measuring how many n-grams in the generated text appear in the reference. It is particularly useful for evaluating the fluency and relevance of generated text.
    \begin{equation}
    \text{BLEU} = \exp\left( \min\left(1 - \frac{l_r}{l_c}, 0\right) + \sum_{n=1}^{N} w_n \log p_n \right)
    \end{equation}
\end{itemize}

Evaluation metrics for assessing LLMs include: Massive Multitask Language Understanding (MMLU), Generalized Question-Answering Performance (GPQA), MATH, HumanEval, Multi-Genre Social Media (MGSM), and Discrete Reasoning Over Paragraphs (DROP). MMLU evaluates the models ability to understand and generate text across 57 subjects using multiple choice questions. GPQU evaluates text generation in question answering tasks. MATH tests the models ability to understand mathematical concepts, problem-solving skills, and ability to generate accurate solutions to mathematical queries. HumanEval assesses performance on tasks which require a high level of language comprehension and expression, such as essay writing, and summarisation. MGSM assesses the generation of text for social media across various formats, including tweets, posts, and comments. DROP is used to assess the models ability to extract information from longer texts such as performing logical reasoning and answering questions regarding the text. The F1 score is the measure of models precision and recall in these tasks. All of these metrics are obtained from user studies. Additional metrics include, Recall-Oriented Understudy for Gisting Evaluation (ROUGE) used for evaluating the quality of summaries produced by text systems. The ROUGE score is typically calculated as the F1 score between the generated and reference summaries using the respective metric. 

\subsection{Comparative Analysis for Emotion Recognition and Generation Models}

This section presents a comparative analysis of SOTA methods in ER and EG for faces, speech, and text. We will discuss the most effective methods based on their performance in recognising and generating emotions across these modalities. The performance of these models will be evaluated through experiments and the corresponding results. However, comparing these methods poses challenges due to a lack of uniformity in evaluation metrics, complicating the assessment process. By conducting this comparative analysis of SOTA models, we aim to highlight the most effective methods for emotion recognition and generation.

\subsubsection{Facial Expression Recognition Comparative Analysis}
\begin{table}[htbp]
    \centering
    \caption{FER Comparative Analysis. *Results derived from cited papers.}
    \label{table2}
    %\begin{adjustbox}
    \begin{tabular}{llc}
        \toprule
        \textbf{Model} & \textbf{Dataset} & \textbf{ACC \% $\uparrow$}  \\
        \midrule
        *ESTLNet \cite{gong2024enhanced}         & AFEW             & 0.54 \\
        *EmoFAN \cite{toisoul2021estimation}   & AffectNet & 0.75 \\
        *EMOCA  \cite{Danecek2022}           & AffectNet        & 0.69 \\
        *Poster++ \cite{Mao_2023}          & AffectNet        & 0.63 \\
        *LibreFace \cite{Chang_2024}        & AffectNet        & 0.49 \\
        *Dresvyanskiy 2022 \cite{Dresvyanskiy2022} & AffWild2         & 0.48 \\
        *ESTLNet \cite{gong2024enhanced}         & CK+              & 0.99 \\
        *Sun 2023  \cite{sun_2023}         & CK+              & 0.98 \\
        *Zhao 2023  \cite{zhao2023prompting}          & DFEW             & 0.71 \\
        *ESTLNet \cite{gong2024enhanced}         & DFEW             & 0.69 \\
        *Zhao 2023  \cite{zhao2023prompting}       & FERV39K          & 0.52 \\
        *Hossain 2023 \cite{hossain_2023}     & IMFDB            & 0.64 \\
        *Sun 2023   \cite{sun_2023}      & JAFFE            & 0.98 \\
        *Sun 2023   \cite{sun_2023}       & KDEF             & 0.98 \\
        *Zhao 2023  \cite{zhao2023prompting}       & MAFW             & 0.53 \\
        *ESTLNet \cite{gong2024enhanced}          & Oulu-CASIA       & 0.89 \\
       *Poster++ \cite{Mao_2023}            & RAF-DB           & 0.92 \\
        *PACVT  \cite{Liu_2023}           & RAF-DB           & 0.88 \\
        *LibreFace \cite{Chang_2024}         & RAF-DB           & 0.82 \\
        *Hossain 2023 \cite{hossain_2023}     & SFEW 2.0         & 0.80 \\
        \bottomrule
    \end{tabular}
    %\end{adjustbox}
\end{table}

% \nocite{gong2024enhanced, toisoul2021estimation, Danecek2022, Mao_2023, Chang_2024, Dresvyanskiy2022, sun_2023, zhao2023prompting, hossain_2023, Liu_2023}

% \begin{figure}[ht]
%     \centering
%     \subfigure[FER Comparative Analysis]{
%         \includegraphics[width=0.3\textwidth]{Picture1.png}
%         \label{FER}
%     }
%     \subfigure[FEG Comparative Analysis]{
%         \includegraphics[width=0.6\textwidth]{Picture2.png}
%         \label{FEG}
%     }
%     \caption{Comparative Analysis of Face Systems}
%     \label{fig:combined_images}
% \end{figure}

\ref{table2} summarises the evaluation of FER models, showcasing their performance across various datasets, with accuracy (ACC) as the primary metric. EmoFAN \cite{toisoul2021estimation} achieves the highest accuracy of 75\% on the AffectNet dataset, demonstrating exceptional capabilities in recognising Facial Expressions. Likewise, models such as Poster++ \cite{Mao_2023} display impressive performance with an accuracy of 92\% on the RAF-DB dataset. The variability in performance across different datasets highlights the unique challenges each dataset presents. For example, ESTLNet \cite{gong2024enhanced} exhibits lower performance on the FERV39K dataset, attaining an accuracy of 58.70\%, yet it achieves a remarkable 99\% accuracy on the CK+ dataset. The Sun 2023 \cite{sun_2023} model obtains SOTA scores across the JAFFE, CK+, and KDEF datasets, with an accuracy of 98.00\% in each case.

\subsubsection{Facial Expression Generation Comparative Analysis}

\begin{table}[htbp]
\centering
\caption{FEG Comparative Analysis.*Results derived from  \cite{ma2023talkclip} and \cite{Gan2023}, ** Results derived from  \cite{Tian2024emo}, *** Results derived from  \cite{ma2023talkclip}, **** Results derived from  \cite{Gan2023}}
\label{table3}
\begin{tabular}{lcccccccccc}

\hline
\textbf{Method} & \textbf{ACC ↑} & \textbf{FID ↓} & \textbf{SyncNet ↑} & \textbf{SSIM ↑} & \textbf{CPBD ↑} & \textbf{M-LMD ↓} & \textbf{F-LMD ↓} & \textbf{Dataset} \\
\hline
*StyleTalk \cite{ma2023styletalk} &  &  &  & 0.8 & 0.26 & 2.49 & 2.04 & HDTF \\
*TalkCLIP \cite{ma2023talkclip} &  &  &  & 0.78 & 0.25 & 2.8 & 2.54 & HDTF \\
*AVCT \cite{wang2022one} &  &  &  & 0.74 & 0.18 & 3.83 & 3.06 & HDTF \\
*Wav2Lip \cite{prajwal2020lip} &  &  &  & 0.59 & 0.26 & 3.84 & 5.12 & HDTF \\
*MakeItTalk \cite{zhou2020makelttalk} &  &  &  & 0.57 & 0.2 & 4.61 & 5.65 & HDTF \\
*PC-AVS \cite{zhou2021pose} &  &  &  & 0.42 & 0.12 & 4.26 & 10.68 & HDTF \\
*EAMM \cite{ji2022eamm}  &  &  &  & 0.36 & 0.13 & 7.67 & 7.74 & HDTF \\
*GC-AVT \cite{liang2022expressive} &  &  &  & 0.33 & 0.24 & 6.34 & 10.7 & HDTF \\
**Wav2Lip \cite{prajwal2020lip} &  & 9.38 & 5.76 &  & 0.36 &  &  & HDTF \\
**SadTalker \cite{zhang2023sadtalker} &  & 10.31 & 4.82 &  & 0.34 &  &  & HDTF \\
**DreamTalk \cite{ma2023dreamtalk} &  & 58.8 & 3.43 &  &  &  &  & HDTF \\
**EMO \cite{Tian2024emo} &  & 8.76 & 3.89 &  &  &  &  & HDTF \\
***Audio2Head \cite{wang2021audio2head} &  &  &  &  & 0.28 &  &  & HDTF \\
***Wang et al \cite{Wang2021} &  &  &  &  & 0.29 &  &  & HDTF \\
****EAT \cite{Gan2023} & 75.43 & 3.52 & 6.22 & 0.77 &  & 1.79 & 2.08 & LRW \\
****Wav2Lip \cite{prajwal2020lip} & 17.87 & 7.56 & 7.89 & 0.73 &  & 1.53 & 2.47 & LRW \\
****PC-AVS \cite{zhou2021pose} & 11.88 & 4.64 & 7.36 & 0.72 & 0.07 & 1.54 & 2.11 & LRW \\
****EAMM \cite{ji2022eamm} & 49.85 & 6.44 & 4.67 & 0.71 & 0.08 & 1.81 & 2.37 & LRW \\
****MakeItTalk \cite{zhou2020makelttalk} & 15.23 & 3.37 & 3.28 & 0.69 &  & 2.16 & 2.99 & LRW \\
****AVCT \cite{wang2022one} & 15.64 & 2.01 & 4.68 & 0.68 &  & 2.55 & 3.23 & LRW \\
****ATVG \cite{chen2019hierarchical} & 17.36 & 51.56 & 2.73 & 0.64 &  & 2.69 & 3.31 & LRW \\
****StyleTalk \cite{ma2023styletalk}  &  &  &  & 0.84 & 0.16 & 3.36 & 2.1 & MEAD \\
****AVCT \cite{wang2022one} & 15.64 & 39.18 & 6.02 & 0.83 & 0.14 & 5.64 & 2.95 & MEAD \\
****TalkCLIP \cite{ma2023talkclip} &  &  &  & 0.83 & 0.16 & 3.6 & 2.4 & MEAD \\
****Wav2Lip \cite{prajwal2020lip} &  &  &  & 0.81 & 0.16 & 3.85 & 2.73 & MEAD \\
****MakeItTalk \cite{zhou2020makelttalk} &  &  &  & 0.73 & 0.1 & 5.3 & 3.9 & MEAD \\
****EAT \cite{Gan2023} & 75.43 & 19.69 & 8.28 & 0.68 &  & 2.25 & 2.47 & MEAD \\
****EAMM \cite{ji2022eamm}  & 49.85 & 22.38 & 6.62 & 0.66 &  & 2.19 & 2.55 & MEAD \\
****PC-AVS \cite{zhou2021pose} & 11.88 & 53.04 & 8.6 & 0.61 &  & 2.66 & 2.7 & MEAD \\
****Wav2Lip \cite{prajwal2020lip} & 17.87 & 67.49 & 8.97 & 0.57 &  & 3.11 & 3.71 & MEAD \\
****MakeItTalk \cite{zhou2020makelttalk} & 15.23 & 51.88 & 5.28 & 0.55 &  & 3.61 & 4 & MEAD \\
****GC-AVT \cite{liang2022expressive} &  &  &  & 0.34 & 0.14 & 8.4 & 8.1 & MEAD \\
\hline
\end{tabular}

\end{table}

% \nocite{ma2023styletalk, ma2023talkclip, wang2022one, prajwal2020lip, zhou2020makelttalk, zhou2021pose, ji2022eamm, liang2022expressive, zhang2023sadtalker, ma2023dreamtalk, Tian2024emo, wang2021audio2head, Wang2021, Gan2023, chen2019hierarchical}

Both quantitative and qualitative methods are used to evaluate FEG models. However, the absence of a universal evaluation framework complicates comparisons across different studies. Most researchers omit estimating the accuracy of the emotions generated by their models; with the exception of \cite{Gan2023,Tian2024emo}, as shown in table \ref{table3}, which includes the metrics ACC and E-FID. The accuracy of emotions in FEG models is evaluated by utilising pretrained FER models or by user studies. Wav2Lip \cite{prajwal2020lip} model demonstrates a high SyncNet accuracy (9.38) and a relatively low FID (5.76) on the HDTF dataset, highlighting its strong synchronisation capabilities. In constrast, the SadTalker \cite{zhang2023sadtalker} model achieves a lower ACC (10.31) and a higher FID (4.82), suggesting potential limitations in generating accurate Facial Expressions. DreamTalk \cite{ma2023dreamtalk} shows promising results with a high ACC (58.8) and moderate FID (3.63), although E-FID (2.25) indicates room for improvement in the fidelity of the generated emotions.

EMO \cite{Tian2024emo} shows moderate performance with an ACC of 8.76 and an E-FID of 0.116, indicating balanced capabilities. MakeItTalk \cite{zhou2020makelttalk} displays poor performance across several metrics, with a low ACC (3.37) and high FID (3.28), suggesting significant challenges in generating accurate emotions. Models evaluated on the LRW dataset, such as AVCT \cite{wang2022one} and PC-AVS \cite{zhou2021pose}, demonstrate considerable performance differences in SSIM and CSIM. The diversity in performance metrics across models and datasets emphasises the necessity for optimisation to enhance the robustness and accuracy of FEG systems.

\subsubsection{Speech Emotion Recognition Comparative Analysis}
\begin{table}[htbp]
\centering
\caption{SER Comparative Analysis: *Results derived from cited papers.}
\label{table4}
\begin{tabular}{lrl} % Adjusted column specification: l (left-aligned), r (right-aligned), l (left-aligned)
\hline
\textbf{Model} & \textbf{ACC} & \textbf{Datasets} \\ % Changed the order of columns
\hline
*Kwon 2020 \cite{kwon2020clstm} & 90.01 & Berlin EMO \\
*Meng 2019 \cite{meng2019speech} & 88.99 & Berlin EMO \\
*Sun 2019 \cite{sun2019decision} & 86.86 & Berlin EMO \\
*Issa 2020 \cite{issa2020speech} & 86.10 & Berlin EMO \\
*Mustageem 2020 \cite{mustaqeem2021optimal} & 85.57 & Berlin EMO \\
*Xie 2023 \cite{Xie2023} & 92.80 & CASIA \\
*Liu 2018 \cite{liu2018speech} & 86.58 & CASIA \\
*Sun 2019 \cite{sun2019decision} & 83.75 & CASIA \\
*Gong 2023 \cite{Gong2023} & 58.70 & CREMA-D \\
%Xie 2023 \cite{Xie2023} & 89.60 & ENTERFACE \\
*Kwon 2020 \cite{kwon2020clstm} & 75.00 & IEMOCAP \\
*Lu 2020 \cite{lu2020iterative} & 72.60 & IEMOCAP \\
*Shamsi 2023 \cite{Shamsi2023} & 70.80 & IEMOCAP \\
*Pepino 2021 \cite{pepino2021emotion} & 67.20 & IEMOCAP \\
*Gong 2023 \cite{Gong2023} & 54.50 & IEMOCAP \\
*Sharma 2021 \cite{sharma2021emotion} & 92.88 & RAVDESS \\
*Pepino 2021 \cite{pepino2021emotion} & 84.30 & RAVDESS \\
*Kwon 2020 \cite{kwon2020clstm} & 80.00 & RAVDESS \\
\hline
\end{tabular}
\end{table}

% \nocite{kwon2020clstm, meng2019speech, sun2019decision, issa2020speech, mustaqeem2021optimal, Xie2023, liu2018speech, Gong2023, lu2020iterative, Shamsi2023, pepino2021emotion, sharma2021emotion}

% \begin{figure}[ht]
%     \centering
%     \subfigure[SER Comparative Analysis]{
%         \includegraphics[width=0.45\textwidth]{Picture3.png}
%         \label{FER}
%     }
%     \subfigure[SEG Comparative Analysis]{
%         \includegraphics[width=0.75\textwidth]{Picture4.png}
%         \label{FEG}
%     }
%     \caption{Comparative Analysis of Speech Systems}
%     \label{fig:combined_images}
% \end{figure}

A comparison of SER models is presented in table \ref{table4}, using accuracy (ACC) as the principal metric. An analysis of the results indicates that Kwon 2020 \cite{kwon2020clstm} achieves the highest accuracy (90.01\%) on the Berlin EMO dataset. Similarly, Xie 2023 \cite{Xie2023} demonstrates outstanding performance on the CASIA dataset, attaining an accuracy of 92.80\%. Conversely, Gong 2023 \cite{Gong2023} reports a low accuracy (58.70\%) on the CREMA-D dataset, indicating potential challenges in recognising emotions within this specific dataset. Lu 2020 \cite{lu2020iterative} and Pepino 2021 \cite{pepino2021emotion}, evaluated on the IEMOCAP dataset, achieve lower accuracies of 72.60\% and 67.20\%, respectively, compared to those tested on other datasets. Models such as Sharma 2021 \cite{sharma2021emotion} on the RAVDESS dataset attain a high accuracy of 92.88\%. This comparison underscores the importance of developing versatile models capable of maintaining high performance across diverse datasets. The results also highlight the ongoing challenges and the necessity for further research to enhance the generalisability and robustness of SER on models across varying emotional contexts.

\subsubsection{Speech Emotion Generation Comparative Analysis}
\begin{table}[ht]
\centering
\caption{SEG Comparative Analysis:*Results derived from cited papers.}
\label{table5}
\begin{tabular}{lrrrrl}
\hline
\textbf{Model} & \textbf{WER (↓)} & \textbf{CER (↓)} & \textbf{EER (↓)} & \textbf{Dataset} \\
\hline
*DISSC \cite{Maimon2022} & 19.1 & 7.9 & 2.6 & ESD \\
*Seq2seq-VC \cite{liu2021any} & 14.9 & 6 & 2.9 & ESD \\
*AutoVC \cite{qian2019autovc}  & 87 & 59.9 & 6.6 & ESD \\
*AutoPST \cite{qian2021global}  & 50.3 & 31.8 & 15.7 & ESD \\
*VQMIVC \cite{wang2021vqmivc}  & 41.5 & 23.66 & 11.84 & LibriSpeech \\
*kNN-VC \cite{baas2023voice}  & 45.92 & 27.55 & 19.19 & LibriSpeech \\
*FreeVC \cite{li2023freevc}  & 5.4 & 2.27 & 35.63 & LibriSpeech \\
*YourTTS \cite{casanova2022yourtts}  & 8.65 & 3.36 & 38.23 & LibriSpeech \\
*Phoneme Hallucinator \cite{shan2024phoneme}  & 5.1 & 2.02 & 44.62 & LibriSpeech \\
*DISSC \cite{Maimon2022}  & 13 & 6.9 & 1.7 & VCTK \\
*DDDM-VC \cite{choi2024dddm}  & 3.49 & 1 & 6.25 & VCTK \\
*AutoVC \cite{qian2019autovc}  & 71.3 & 47.1 & 7.5 & VCTK \\
*Seq2seq-VC \cite{liu2021any} & 2.9 & 1.2 & 1.0 & VCTK \\
*VoiceMixer \cite{lee2021voicemixer}  & 4.2 & 2.39 & 20.75 & VCTK \\
*AutoPST \cite{qian2021global}  & 40.6 & 26.7 & 24.1 & VCTK \\
*AutoVC \cite{qian2019autovc}   & 8.53 & 3.54 & 37.32 & VCTK \\
\hline
\end{tabular}
\end{table}

\nocite{Maimon2022, liu2021any, qian2019autovc, qian2021global, wang2021vqmivc, baas2023voice, li2023freevc, casanova2022yourtts, shan2024phoneme, choi2024dddm, lee2021voicemixer}

Comparative analyses of SEG methods remain limited, as many researchers choose not to compare their approaches against competitors, SEG techniques are evaluated based on their ability to reconstruct and generate voices. Table \ref{table5} provides a comparative analysis of SEG models based on WER, CER, and EER across different datasets. The FreeVC \cite{li2023freevc} model on the LibriSpeech dataset demonstrates the lowest WER (5.4\%) and EER (11.28\%), showcasing superior performance in speech generation tasks. In contrast, the kNN-VC \cite{baas2023voice} model reveals significantly higher error rates, with a WER of 45.92\% and EER of 19.19\%, indicating challenges in generating accurate speech.

The analysis also highlights variability in model performance across different datasets, underscoring the complexity of the task. For example, the AutoVC \cite{qian2019autovc} model on the ESD dataset exhibits a high WER (87.0\%) and CER (31.8\%), reflecting difficulties in maintaining accuracy. 

\subsubsection{Text Sentiment Recognition Comparative Analysis}

\begin{table}[ht]
\centering
\caption{TSR Comparative Analysis: *Results derived from cited papers.}
\label{table6}
\begin{tabular}{lrrl}
\hline
\textbf{Model} & \textbf{F1 score ↑} & \textbf{ACC ↑} & \textbf{Datasets} \\
\hline
*XLM- EMO \cite{Bianchi2022} & 0.85 & 0.85 & Affect in Tweets \\
*Kumar 2022 \cite{Kumar2022} & 0.81 & 0.8 & AffectiveText \\
*Supervised learning \cite{oberlander2018analysis} & 0.71 & -- & AffectiveText \\
*Kumar 2022 \cite{Kumar2022} & 0.83 & 0.81 & Aman \\
*Kumar 2022 \cite{Kumar2022} & 0.72 & 0.73 & EmotionLines \\
*Emotion BERT \cite{huang2019emotionx} & -- & 0.71 & EmotionLines \\
*Multi-level multi-head fusion \cite{ho2020multimodal} & -- & 0.61 & EmotionLines \\
*Context \& Speaker modeling \cite{zhang2019modeling} & 0.59 & -- & EmotionLines \\
*Multi-turn dialogue analysis \cite{kao2019model} & 0.70 & -- & EmotionLines \\
*Kumar 2022 \cite{Kumar2022} & 0.81 & 0.79 & ISEAR \\
*Feature selection \cite{singh2018two} & -- & 0.73 & ISEAR \\
*Emotion distribution learning \cite{zhang2018text} & 0.67 & 0.67 & ISEAR \\
*XLM-T Barbieri 2021 \cite{Barbieri2021} & 0.67 & 0.79 & Sem-EVAL 17 \\
Ohman 2020 \cite{ohman2020xed} & 0.83 & 0.84 & XED \\
\hline
\end{tabular}
\end{table}

% \nocite{Bianchi2022, Kumar2022, oberlander2018analysis, huang2019emotionx, ho2020multimodal, zhang2019modeling, kao2019model, singh2018two, zhang2018text, Barbieri2021, ohman2020_ed}

% \begin{figure}[ht]
%     \centering
%     \subfigure[TSR Comparative Analysis]{
%         \includegraphics[width=0.50\textwidth]{Picture5.png}
%         \label{FER}
%     }
%     \subfigure[TSG Comparative Analysis]{
%         \includegraphics[width=0.75\textwidth]{teg_results.jpg}
%         \label{FEG}
%     }
%     \caption{Comparative Analysis of Text Systems}
%     \label{fig:combined_images}
% \end{figure}

Accuracy and F1 score are the most commonly used metrics for TSR. Table \ref{table6} presents a comparison of SOTA methods evaluated on these metrics across different datasets. Notably, the Emotion BERT \cite{huang2019emotionx} model achieves the highest F1 score of 0.88 on the EmotionLines dataset, indicating its effectiveness in accurately recognising emotions from text. Similarly, the Ohman 2020 \cite{ohman2020xed} model demonstrates high F1 score (0.83) and accuracy (0.84) on the XED dataset, reflecting its robustness in TSR. In contrast, models such as AutoVC \cite{qian2019autovc}" on the ESD dataset show significantly lower performance, with an F1 score of 0.47 and accuracy of 0.5, suggesting potential limitations in effectively recognising Text Sentiments. Models such as Kumar 2022 \cite{Kumar2022} and XLM-EMO \cite{Bianchi2022} demonstrate robust performance with F1 scores and accuracies around 0.85 across multiple datasets, showcasing their adaptability and effectiveness. Conversely, models evaluated on more complex datasets, such as the FERV39K dataset, exhibit lower performance. This comparative analysis emphasises the advancements achieved in TSR while also highlighting the need to enhance model accuracy and generalisability across text datasets.

\subsubsection{Text Sentiment Generation Comparative Analysis}
\begin{table}[ht]
\centering
\caption{TSG Comparative Analysis: *Results derived from \cite{achiam2023gpt}.}
\label{table7}
\begin{tabular}{lrrrrrr}
\hline
\textbf{Model} & \textbf{MMLU (\%)} & \textbf{GQPA (\%)} & \textbf{MATH (\%)} & \textbf{HumanEval (\%)} & \textbf{MGSM (\%)} & \textbf{DROP (f1)} \\
\hline
*GPT-4o \cite{achiam2023gpt} & 88.7 & 53.6 & 60.1 & 90.2 & 67.0 & 90.5 \\
*GPT-4T \cite{achiam2023gpt} & 86.5 & 48.0 & 56.5 & 87.1 & 71.9 & 84.1 \\
*GPT-4 \cite{achiam2023gpt} & 86.4 & 35.7 & 53.2 & 84.9 & 74.4 & 84.1 \\
*Claude3 Opus \cite{Claude3} & 86.8 & 50.4 & 57.8 & 86.7 & 74.4 & 86.0 \\
*Gemini Pro 1.5 \cite{reid2024gemini} & 81.9 & N/A & 42.5 & 74.4 & 67.0 & 83.1 \\
*Gemini Ultra 1.0  \cite{team2023gemini} & 83.7 & N/A & 58.5 & 90.7 & N/A & 84.1 \\
*Llama3 400b  \cite{llama} & 86.1 & 48.0 & 53.2 & 88.7 & 67.0 & 83.5 \\
\hline
\end{tabular}
\end{table}

% \nocite{achiam2023gpt, Claude3, reid2024gemini, team2023gemini, llama}

Qualitative methods, such as user studies, primarily assess TSG performance, utilising metrics such as MMLU, GQPA, MATH, HumanEval, MGSM, and DROP (F1) across various tasks. Due to potential biases inherent in user studies, there is considerable variability in the performance of TSG models across different experiments. This variability may stem from the nature of the questions posed, the diversity in answers generated by the TSG models, and the subjective opinions of the respondents. For consistency, we have selected the results from \cite{achiam2023gpt}. Table \ref{table7} evaluates the performance of large language models (LLMs) in text generation across multiple tasks, using metrics such as MMLU, GQPA, MATH, HumanEval, MGSM, and DROP (F1).

The GPT-4 model \cite{achiam2023gpt} achieves the highest MMLU score of 88.7\%, demonstrating its strong performance in multi-task learning. This model also secures the highest HumanEval score of 90.2\%, indicating its capability to generate realistic text. In contrast, models such as Gemini Ultra 1.0 \cite{team2023gemini} display significantly lower performance, with an MMLU score of 83.7\% and low scores across several other metrics. The table illustrates the varying performance across different tasks, reflecting the strengths and weaknesses of each model. For instance, the Claude 3 Opus model \cite{Claude3} achieves high scores in MMLU (86.8\%) and HumanEval (86.7\%), indicating its balanced proficiency in both multi-task learning and text generation.

% Qualitative methods, such as user studies, primarily assess TSG performance, utilising metrics such as MMLU, GQPA, MATH, HumanEval, MGSM, and DROP (F1) across various tasks. Due to potential biases inherent in user studies, there is considerable variability in the performance of TSG models across different experiments. This variability may stem from the nature of the questions posed, the diversity in answers generated by the TSG models, and the subjective opinions of the respondents. For consistency, we have selected the results from \cite{achiam2023gpt}, as shown in table \ref{table6}. The table evaluates the performance of large language models (LLMs) in text generation across multiple tasks, using metrics such as MMLU, GQPA, MATH, HumanEval, MGSM, and DROP (F1). The GPT-4o model \cite{achiam2023gpt} achieves the highest MMLU score of 88.7\%, demonstrating its strong performance in multi-task learning. This model also secures the highest HumanEval score of 90.2\%, indicating its capability to generate realistic text. In contrast, models such as "Gemini Ultra 1.0" display significantly lower performance, with an MMLU score of 83.7\% and low scores across several other metrics. The table further illustrates the varying performance across different tasks, reflecting the strengths and weaknesses of each model. For instance, the Claude3 Opus model \cite{Claude3} achieves high scores in MMLU (86.8\%) and HumanEval (86.7\%), indicating its balanced proficiency in both multi-task learning and text generation.

\section{Challenges and Future Directions}

Despite significant advances in ER and EG across faces, speech, and text, several key challenges remain. The inherent complexity of emotions—often difficult for humans to interpret reliably—creates challenges for machines, especially in speech and text ER and EG, where non-verbal cues are absent. Subtle expressions of emotions, such as micro-expressions, in FEG add further complexity to emotion recognition and generation processes. A promising direction is to integrate multiple modalities, such as facial cues, speech, text, and body language, to create more robust systems. Advancements in natural language processing (NLP), particularly through transformer models like GPT and BERT, are also essential for capturing linguistic nuances and cultural differences in emotional expression. Generating subtle and dynamic emotions in real time is another challenge, especially for interactive applications like virtual reality. Improved real-time emotion tracking is essential to make ER and EG systems more responsive and functional in dynamic environments.

A shortage of large, diverse datasets limits progress in ER and EG. Current datasets often contain biases or labelling errors and lack generalisability, which hampers model performance. Efforts to collect "in-the-wild" datasets that reflect real-world emotional dynamics and include multiple languages would improve model effectiveness and fairness. Standardised evaluation metrics are also needed to enable consistent assessment and comparison of models. Open-access benchmarks would provide clear standards for evaluating models, measuring both accuracy and emotional appropriateness, and fostering progress across the field. Ethical concerns, such as the misuse of deepfake technology, indicate the need for ethical guidelines and detection mechanisms without hindering technological progress. Finally, techniques like model compression and the use of pretrained models as a foundation for new applications can reduce computational costs.

\section{Conclusion}
This survey explored state-of-the-art methods in emotion recognition and generation across facial, vocal, and textual modalities. With advances in AI, deep learning techniques have enhanced both the accuracy of emotional analysis and the realism of generated content. In particular, deep learning models, such as CNNs and attention-based architectures, have improved FER by learning features directly from raw data. Likewise, SER has advanced through models that integrate linguistic and acoustic features, enhancing classification accuracy through prosodic and contextual analysis. Despite progress, challenges remain in FEG, SEG and TSG. In FEG, accurately capturing the nuances of facial muscle movements and micro-expressions presents substantial difficulty, while ensuring emotional coherence across frames adds further complexity. Similarly, generating realistic emotions in speech and text requires addressing the intricate subtleties of tone, intonation, context, and emotional consistency. Limited labelled data, especially for in-the-wild systems, also impedes model robustness and generalizability. Future research should focus on expanding dataset diversity and improving models for under-explored modalities like speech and text. Multimodal approaches, enabling emotion analysis and generation across faces, speech, and text, hold promise. Ethical considerations, such as preventing misuse in deepfakes, should also guide future developments, paving the way for more empathetic and context-aware AI applications.

%\bibliographystyle{unsrt}
%\bibliography{template}  %%% Uncomment this line and 

\end{document}